\documentclass{article}
\usepackage[T1]{fontenc}
\PassOptionsToPackage{numbers,compress}{natbib}
\usepackage[dblblindworkshop, final]{neurips_2025}

\usepackage{microtype}
\usepackage{graphicx}
\usepackage{subcaption}
\usepackage{booktabs} 
\usepackage{amsmath,amssymb}
\usepackage{hyperref}
\usepackage{url}
\usepackage{xcolor}

\usepackage{listings}
\lstdefinelanguage{JavaScript}{
  keywords={break, case, catch, continue, debugger, default, delete, do, else, finally, for, function, if, in, instanceof, new, return, switch, this, throw, try, typeof, var, void, while, with, const, let},
  keywordstyle=\color{blue}\bfseries,
  ndkeywords={class, export, boolean, throw, implements, import, this},
  ndkeywordstyle=\color{darkgray}\bfseries,
  identifierstyle=\color{black},
  sensitive=false,
  comment=[l]{//},
  morecomment=[s]{/*}{*/},
  commentstyle=\color{gray}\ttfamily,
  stringstyle=\color{red}\ttfamily,
  morestring=[b]',
  morestring=[b]"
}
\usepackage[most]{tcolorbox}
\lstdefinestyle{duolens}{
  basicstyle=\ttfamily\small,
  numbers=left,
  numberstyle=\tiny,
  stepnumber=1,
  numbersep=6pt,
  tabsize=2,
  showstringspaces=false,
  breaklines=true,
  frame=tb,
  framerule=0.4pt,
  rulecolor=\color{black!15},
  keywordstyle=\color{blue!70!black},
  commentstyle=\itshape\color{teal!70!black},
  stringstyle=\color{magenta!50!black}
}
\newtcolorbox{promptbox}[1][]{enhanced,breakable,sharp corners,
  colback=blue!2,colframe=blue!50!black,boxrule=0.5pt,title=#1}

\hypersetup{colorlinks=true, linkcolor=blue, citecolor=blue, urlcolor=blue}

\title{DuoLens: A Framework for Robust Detection of Machine-Generated Multilingual Text and Code}

\author{%
  Shriyansh Agrawal$^{*}$\\
  \texttt{shriyanshag0411@gmail.com} \\
  \And
  Aidan Lau$^{*}$\\
  \texttt{aidan3616@outlook.com} \\
  \And
  Sanyam Shah\\
  \texttt{sanyam21in@gmail.com} \\
  \And
  Sunishchal Dev\\
  \texttt{dev@algoverseairesearch.org} \\
  \And
  Vasu Sharma\\
  \texttt{sharma.vasu55@gmail.com} \\
  \And
  Kevin Zhu\\
  \texttt{kevin@algoverseacademy.com} \\
  \And
  Ahan M R$^{\dagger}$\\
  \texttt{ahanmr@microsoft.com} \\
  \\[0.5em]
}

\begin{document}
\maketitle

\renewcommand{\thefootnote}{*}
\footnotetext{Equal contribution}
\renewcommand{\thefootnote}{$\dagger$}
\footnotetext{Project lead}
\renewcommand{\thefootnote}{\arabic{footnote}}

\begin{abstract}
The prevalence of Large Language Models (LLMs) for generating multilingual text and source code has only increased the imperative for machine-generated content detectors to be accurate and efficient across domains. Current detectors either incur high computational cost or lack sufficient accuracy, often with a trade-off between the two, leaving room for further improvement. To address these gaps, we propose the fine-tuning of encoder-only Small Language Models (SLMs), in particular, the pre-trained models of RoBERTA \&  CodeBERTa using specialized datasets on source code and other natural language to prove that for the task of binary classification, SLMs outperform LLMs by a huge margin whilst using a fraction of compute. Our encoders achieve AUROC = 0.97 to 0.99 and macro-F1 = 0.89 to 0.94 while reducing latency by $8$-$12\times$ and peak VRAM by $3$--$5\times$ at 512-token inputs.
Under cross-generator shifts and adversarial transformations (paraphrase, back-translation; code formatting/renaming), performance retains \textbf{$\ge 92\%$} of clean AUROC.
We release training and evaluation scripts with seeds and configs; a reproducibility checklist is also included.
\end{abstract}

\section{Introduction}
As generative AI models continue to advance, their outputs permeate many domains, from written prose to computer code. AI-assisted content creation can boost productivity---for instance, code generation with LLMs has been reported to improve developer efficiency by up to 33\% \citep{becker2022programminghardused}. However, the rise of machine-generated text and code also brings critical concerns in academia and industry, such as plagiarism, misinformation, and unfair advantages in assessments or job interviews questioned by \citet{academic}. Furthermore, this issue extends toward source code, exacerbated by the wide availability of specialized LLMs (Qwen Coder, Anthropic Claude 3.7/4 Sonnet) and coding 'Agents' which are capable in acting autonomously in a workflow. They exhibit a track record for inefficient code generation, poor security practices, partial execution and the long running issue of LLM hallucination from \citet{security}. This has sparked a growing need for reliable and accessible detection of machine-generated content to maintain integrity and trust. Current detectors often exhibit one notable weaknesses: many are primarily effective only on English text and struggle with other languages, leading to false positives on non-English inputs. Additionally, individual detection approaches (e.g., based on perplexity or stylistic cues alone) can be evaded, and ensemble agent-based methods can be prohibitively slow or resource-intensive, such as  \citet{li2025agentxadaptiveguidelinebasedexpert}'s method. Similar challenges apply to source code detection---detectors must distinguish machine-generated code from human-written code across different programming languages, but research in this area is quite narrow but readily emerging  \citep{suh2024empiricalstudyautomaticallydetecting}. 

In this work, we propose the use of SLMs rather than LLMs for binary classification of code into machine-generated or human-written labels. Our rationale for the use of SLMs  is since that: 1) LLMs may have a propensity or underlying bias to viewing AI generated text as being of higher quality; 2) LLMs are prone to hallucinate and, when fine-tuned, may experience catastrophic failure. 


\paragraph{Contributions:} 
\begin{itemize}
    \item \textbf{Datasets for Evaluating Detection in Source Code and Multilingual Text:} We present a large-scale dataset that aligns well with the optimal input range of SLMs (with chunking employed when necessary). The datasets encompass samples of source code spanning multiple programming languages, as well as multilingual text from diverse domains, all annotated with binary classification labels.
    \item \textbf{Comprehensive Evaluation:} We conduct benchmarking experiments on multilingual text, as well as source code in Python, Java, JavaScript, C, C++, C\#, and Go. Across these evaluations, DuoLens consistently outperforms baseline detectors in terms of accuracy and macro-F1, including under challenging scenarios such as cross-model and cross-language settings. These findings support our hypothesis that SLMs can surpass LLMs in the binary classification of human-written versus machine-generated code and multilingual text.
\end{itemize}
\setlength{\parskip}{0pt}
\section{Related Work}
\paragraph{Pre-existing Datasets.} 
Although the field remains relatively nascent, a number of datasets have been developed to distinguish between human-written and machine-generated code. \textbf{AIGCodeSet} \citep{aigcd} consists exclusively of Python samples, exemplifying a wider trend also observed in \textbf{CodeMirage} \citep{codemirage} and \textbf{DroidCollection} \citep{droidcollect}, where Python constitutes the majority of instances while other languages (C, C++, C\#, Java, JavaScript, Go) have a sparse representation. These resources also incorporate task-specific samples and enhance stylistic diversity by including outputs from multiple LLMs. In contrast, resources for multilingual text detection exhibit greater variation. \textbf{Multitude} \citep{multitude} compiles news articles but lacks syntactic alignment, while \textbf{Multisocial} \citep{multisocial} captures authentic social media content yet is limited to informal registers. \textbf{M4} \citep{m4} offers a substantially larger corpus, albeit predominantly formal in style. Finally, \textbf{HC3} \citep{hc3} provides multilingual QA pairs but remains constrained in its linguistic coverage.
\paragraph{RoBERTa as a detector.} 
Early detector efforts fine-tuned RoBERTa on model-specific outputs (e.g., OpenAI’s RoBERTa-based GPT-2 detector from \citet{gpt2}; community releases such as “roberta-large-openai-detector”), demonstrating high accuracy on the generator(s) seen during training but poor generalization to unseen or larger models. For example, \citet{gptsentinel}’s GPT-Sentinel, had similar results where it was unable to generalise to other LLMs. However, recent research has moved away from enhancing or enriching RoBERTa and BERT models in favor of using LLMs for text detection, which includes methods such as Binoculars \citep{hans2024spottingllmsbinocularszeroshot}, AGENT-X \citep{li2025agentxadaptiveguidelinebasedexpert}, Ghostbuster \citep{verma2024ghostbusterdetectingtextghostwritten}, EAGLE \citep{bhattacharjee2024eagledomaingeneralizationframework}, and RAIDAR \citep{Mao2024RaidarGA}. However, we advocate for the use of BERT-based models, particularly for binary classification tasks, where LLMs demand substantially greater computational resources while yielding inferior or, at best, comparable performance.

\begin{figure}
    \centering
    \includegraphics[width=1\linewidth]{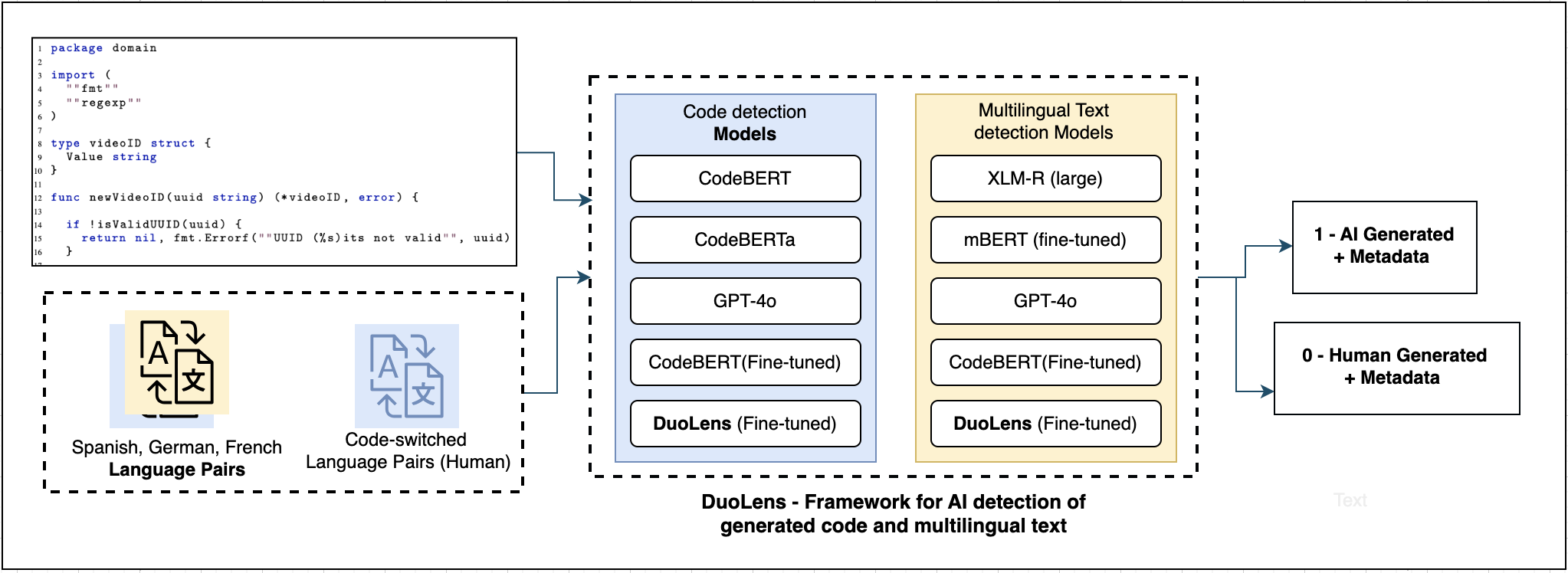}
    \caption{Overview of the experimental setup where each sample from the dataset is processed by the models and classified as either 0 (human-written) or 1 (machine-generated)}
    \label{fig:expsetup}
\end{figure}

\section{Methodology}
\paragraph{Multilingual Dataset.} Many prior multilingual datasets are heavily topic-focused, for example, concentrating on singular domains such as social media or news, which limits the models ability to generalize across different styles of writing. Additionally, they also contain heavy language imbalances (most towards English), which leads to worse performance on other languages. 

To address these issues, our dataset is comprised of 54,520 text samples across eight languages: English, Russian, Arabic, Dutch, German, Spanish, Portuguese, and Romanian, achieved through extending previous datasets, including \textbf{MULTITuDE} \citep{multitude}  (news focused), \textbf{MultiSocial} \citep{multisocial} (social media focused), \textbf{M4} \citep{m4} (general but slightly outdated), and \textbf{HC3} (question-answer pair focused). Additional details about the dataset are available in Appendix \ref{subsec:multilingual}.
\setlength{\parskip}{0pt} Our dataset is more balanced in terms of style and domain, making it better suited for fine-tuning. Similarly to the code dataset, it has an equal balance of human-written and machine-generated samples (27,260 samples for each) to prevent any bias from arising. However, the languages are not balanced in this dataset since the datasets used to create ours all had different languages with varying amounts of samples in shared languages between them, leading to language imbalances in our dataset. 
\paragraph{Code Dataset.} Existing datasets for code generation evaluation exhibit two notable limitations: \textbf{1)} they are often narrow in scope, with samples predominantly drawn from sources such as \citet{codenet}'s IBM CodeNet or competitive programming platforms thereby offering limited representation of production-level code; \textbf{2)} they tend to be heavily imbalanced across languages, with a disproportionate emphasis on Python, which hinders the ability to generalize across syntactically complex languages such as the C family of languages as well as Java, and JavaScript.
Our dataset was constructed by curating samples from previous datasets, including \textbf{AIGCD} \citep{aigcd}, \textbf{DroidCollection} \citep{droidcollect}, and \textbf{CodeMirage} \citep{codemirage}. To solve the aforementioned problems, our dataset contains code samples in 7 programming languages: Python, Java, JavaScript, C, C\#, C++, and Go. Each language has an identical number of human-written and machine-generated samples (6000 samples for each class in its respective language), to ensure balance not only between the classes but also between the languages, in turn solving both the problems that hindered previous datasets. Shown in Figure \ref{fig:langdistro} and explained in Appendix \ref{subsec:sourcecode}.
\setlength{\parskip}{0pt}
\paragraph{Dataset Creation} Both datasets were constructed by first enumerating the available samples for each label (machine-generated and human-written) within each language, in order to determine the maximum number of samples that could be retained per language while preserving label balance. This procedure ensured that no label or language was overrepresented, thereby mitigating potential sources of bias that could adversely affect fine-tuning. Once these maxima were established, the final samples were drawn from the aggregated pool of all available datasets for the particular dataset (multilingual or source code).
\setlength{\parskip}{0pt}
\paragraph{Architecture.} For the task of detecting machine-generated code, we employed \textbf{CodeBERT} by \citet{codebert} and \textbf{CodeBERTa} as baseline models. CodeBERT is trained on natural language and source code, thereby possessing capabilities in both domains. In contrast, CodeBERTa is half the size of CodeBERT and is exclusively trained on source code (from CodeSearchNet \citet{codesearchnet}) without official support for NLP tasks. Both models were fine-tuned through a classification head attached to each model at initialization that was subsequently trained and fine-tuned, using samples from all the languages in the dataset, to distinguish between machine-generated and human-written code. We propose \textbf{DuoLens}, a dual-encoder detector that fuses complementary representations from \textbf{CodeBERT} and \textbf{CodeBERTa} to identify AI-generated content in both natural language and source code. Given an input sequence x, we obtain hidden states from each encoder and derive pooled vectors via \textbf{[CLS]} or mean pooling; a lightweight fusion head then combines these signals using a learned gate that down-weights redundant features and emphasizes encoder-specific cues (e.g., lexical/semantic alignment from CodeBERT’s NL–code pretraining versus syntactic/structural regularities from CodeBERTa’s code-centric pretraining). The fused representation feeds a single linear classifier trained with class-balanced binary cross-entropy.
\setlength{\parskip}{0pt}
\paragraph{Multilingual encoders.}
For multilingual text classification, we fine-tune three widely used pretrained encoders: \textbf{XLM-RoBERTa-base} and \textbf{XLM-RoBERTa-large} \citep{xlmroberta}, and \textbf{Multilingual BERT} (mBERT) \citep{mbert}. 
The two XLM-RoBERTa variants are trained on the same CommonCrawl-derived multilingual corpus and differ only in model capacity, whereas mBERT is trained primarily on multilingual Wikipedia. 
All models are fine-tuned using the protocol described above with identical tokenization, maximum sequence length, optimizer, and learning-rate schedule; checkpoints are selected on the development set, and probabilities are calibrated with temperature scaling.
\setlength{\parskip}{0pt}

\section{Experiments}
\subsection{Experimental Setup}

\paragraph{Baselines.} We evaluated the base models of \textbf{CodeBERT} and \textbf{CodeBERTa} using a probing approach described in \citet{tenney2019bertrediscoversclassicalnlp}'s work. An analogous procedure was employed for the non-fine-tuned multilingual models. We also employed a chunking strategy for the classification of source code, as many samples exceeded the maximum input length of 512 tokens for BERT-based models. In addition, we incorporated \textbf{GPT 4o} (\citet{gpt4o}) as a baseline for both domains and \textbf{Qwen2.5 Coder 3B} for source code (\citet{hui2024qwen2}), which was assessed through few-shot prompting with six examples (3 pairs of machine-generated and human-written samples in different languages). We choose to use this prompt engineering method as it reflects a more realistic scenario which then better facilitates reproducibility and further work. 

\paragraph{Metrics.} We report overall accuracy (class and language specific as well), AUCROC, F1-Macro, and samples per second for all models. 

\paragraph{Scenarios.} We evaluate: (1) Detection of source code in all the 7 programming languages included in our dataset (2) Detection of multilingual text in the 8 languages provided in our dataset (3)  Cross language performance for language specific fine-tuned checkpoints of the model. 
\setlength{\parskip}{0pt}
\subsection{Results and Analysis}

\paragraph{Multilingual Text Detection.}Table \ref{tab:detection_results} presents the performance of the models on multilingual text detection. Similar to the code domain, GPT-4o underperforms with an AUCROC of 0.573 and F1-Macro of 0.490, demonstrating limited binary classification capabilities in this specific setting. Pretrained multilingual models, especially XLM-RoBERTa-large, show reasonably strong performance out of the box, with AUCROC values ranging from 0.891 to 0.937. 

When fine-tuned, all the SLMs achieve very similar performance among each other, with XLM-RoBERTa-large achieving an especially high F1-Macro score, which could be explained due to its much larger size. Additionally, the results from the cross-language scenario are included in Table \ref{tab:multiCrossLang}

\paragraph{Code Detection.}Table \ref{tab:detection_results} showcases the performance of models on the task of detecting machine-generated code. The baseline models include CodeBERT \citep{codebert}, CodeBERTa, and GPT-4o \citep{gpt4o}, with their respective fine-tuned variants. Particularly, GPT-4o, a powerful LLM, performs significantly worse than SLMs on both metrics, achieving only 0.535 AUCROC and 0.414 F1-Macro. In contrast, CodeBERT and CodeBERTa, even without fine-tuning, achieve substantially higher AUCROC scores of 0.953 and 0.948, respectively, all while requiring substantially less compute.

Upon fine-tuning, both CodeBERT and CodeBERTa achieve near-perfect performance. CodeBERTa (fine-tuned) achieves very similar results to CodeBERT (while being half the size of CodeBERT) with an AUCROC of 0.985 and F1-Macro of 0.937, surpassing CodeBERT but(fine-tuned) not by much.

Additionally, the results from the cross-language performance scenario are included in Table \ref{tab:crosslang}





\begin{table}[t]
\centering
\caption{Detection performance: (i) \emph{Code detection} and (ii) \emph{Multilingual text detection} reported as AUC-ROC and F1-Macro. Highest score per column is in \textbf{bold}.}
\label{tab:detection_results}
\begin{minipage}{0.48\linewidth}
\centering
\small
\begin{tabular}{lcc}
\toprule
\textbf{Model} & \textbf{AUC-ROC} & \textbf{F1-Macro} \\
\midrule
CodeBERT & 0.953 & 0.888 \\
CodeBERTa & 0.948 & 0.883 \\
Qwen2.5 Coder 3B & 0.492 & 0.388 \\
GPT-4o & 0.535 & 0.414 \\
CodeBERT (fine-tuned) & 0.98 & 0.93 \\
\textbf{DuoLens (fine-tuned)} & \textbf{0.985} & \textbf{0.937} \\
\bottomrule
\end{tabular}

\medskip
(i) Code detection
\end{minipage}\hfill
\begin{minipage}{0.48\linewidth}
\centering
\small
\begin{tabular}{lcc}
\toprule
\textbf{Model} & \textbf{AUC-ROC} & \textbf{F1-Macro} \\
\midrule
XLM-R-large & 0.937 & 0.865 \\
XLM-R & 0.915 & 0.836 \\
mBERT & 0.891 & 0.816 \\
GPT-4o (few-shot) & 0.573 & 0.49 \\
XLM-R-large (fine-tuned) & 0.974& \textbf{0.924} \\
XLM-R (fine-tuned) & 0.974 & 0.899 \\
\textbf{DuoLens (fine-tuned)} & \textbf{0.975 }& 0.911 \\
\bottomrule
\end{tabular}

\medskip
(ii) Multilingual text detection
\end{minipage}
\end{table}

\section{Conclusion}
We presented \textbf{DuoLens}, a dual-encoder system that unifies detection of AI-generated text and code via the combination of CodeBERT \citep{codebert}and CodeBERTa using a fine-tuned fusion head. DuoLens outperforms strong baselines on multilingual text and source code across the languages in our datasets, and shows improved cross-language generalization (specifically, in Java).  Further details regarding directions for future work are included in Appendix \ref{subsec:futurework}. By building upon the  approach of fine-tuning BERT models, but with a keen view towards effective and balanced dataset creation, we then effectively address the pertinent issue of detecting AI generated content across domains -- making this more accessible regardless of compute limitations or available models. 

\section{Limitations}
At this stage, DuoLens focuses on the use of SLMs in binary classification of AI-generated natural language across modalities. However, due to the inherent limitation of BERT based models being encoder-only, there is no output visible to the user.  This potentially would leave room for future work, in which an LLM could be integrated for sentence-level classification. The datasets we created can also be improved, specifically the multilingual text dataset which has language imbalances as illustrated in Figure \ref{fig:multiLangDistro}. Additionally, DuoLens inherits biases of its underlying models; while we try our best to train on diverse data, fairness is not guaranteed. A similar issue is apparent where even with robust evaluation, results may be constrained to specific languages and models within their respective training and test sets. 

Furthermore, we must also acknowledge that due to limited computational resources and for the purpose of efficacy our baselines are largely constrained to open-weight LLMs, with only GPT 4o used as a closed source AI generator. 

\bibliographystyle{plainnat}
\bibliography{sources}

\newpage

\appendix
\section*{Appendix: Code Samples, Prompt Templates, and Multi-LLM Protocol}

\begin{figure}
    \centering
    \includegraphics[width=0.75\linewidth]{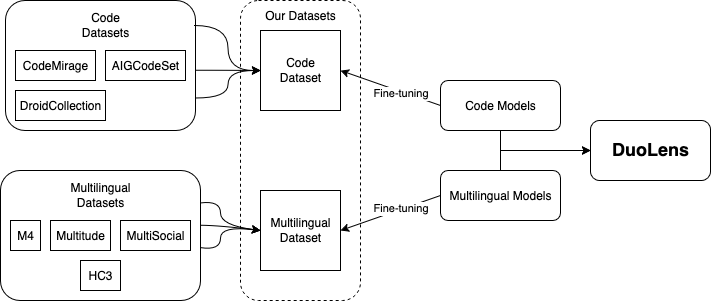}
    \caption{An overview of the methodology adopted in this work is as follows. We first constructed new datasets by curating and extending samples from existing resources. These datasets were subsequently employed to fine-tune and evaluate the models that constitute DuoLens.}
    \label{fig:methodology}
\end{figure}

This appendix provides representative code samples (human-written vs.\ AI-generated) for each programming language (Python, Java, Go, C++, and JavaScript). Additionally, an overview of the methodology is presented in Figure \ref{fig:methodology}

\section{Code Samples from our Dataset}
\subsection{Python Samples}
\paragraph{Human-Written}
\begin{lstlisting}[style=duolens,language=Python]
N=int(input())
Z=[0]*N
W=[0]*N

for i in range(N):
    x,y=map(int,input().split())
    Z[i]=x+y
    W[i]=x-y

alpha=max(Z)-min(Z)
beta=max(W)-min(W)
print(max(alpha,beta))
\end{lstlisting}

\paragraph{AI-Generated}
\begin{lstlisting}[style=duolens,language=Python]
def flatten(m, p=()):
    """"""
    Flattens a mapping tree so that all leaf nodes appear 
    as tuples in a list containing a path and a value.
    
    Parameters:
    m (dict): A dictionary that may contain other dictionaries.
    
    Returns:
    list of tuples: A list of tuples where the first item is 
                    a tuple representing the path to the leaf node
                    and the second item is the value of the leaf node.
    """"""
    result = []
    for k, v in m.items():
        if isinstance(v, dict):
            result.extend(flatten(v, p + (k,)))
        else:
            result.append((p + (k,), v))
    return result

\end{lstlisting}

\subsection{Java Samples}
\paragraph{Human-Written}
\begin{lstlisting}[style=duolens,language=Java]
private DataList fillDataList(List<Global> results, long records, Query query, Map<String, QueryParameter> parameterMap) throws AWException {
    DataListBuilder builder = getBean(DataListBuilder.class);
    boolean paginate = query == null || !query.isPaginationManaged();
    builder.setEnumQueryResult(results)
      .setRecords(records)
      .setPage(parameterMap.get(AweConstants.QUERY_PAGE).getValue().asLong())
      .setMax(parameterMap.get(AweConstants.QUERY_MAX).getValue().asLong())
      .paginate(paginate)
      .generateIdentifiers();

    // If query is defined, fill with query data
    if (query != null) {
      // Add transformations & translations
      builder = processDataList(builder, query, parameterMap);
    }

    // Sort datalist
    builder = sortDataList(builder, parameterMap);

    return builder.build();
  }
\end{lstlisting}

\paragraph{AI-Generated}
\begin{lstlisting}[style=duolens,language=Java]
public void scheduleDelayedTask() {
    mLoginStarted = false;
    Handler handler = new Handler(Looper.getMainLooper());
    Runnable task = new Runnable() {
        @Override
        public void run() {
            // Execute the task
            // You can add any code here that needs to run after the delay
        }
    };
    
    Log.d(""TaskScheduler"", ""Scheduling task with a delay of "" + WAIT_FOR_LOGIN_START_MS + "" milliseconds."");
    handler.postDelayed(task, WAIT_FOR_LOGIN_START_MS);
}
\end{lstlisting}

\subsection{Go Samples}
\paragraph{Human-Written}
\begin{lstlisting}[style=duolens,language=Go]
package domain

import (
	""fmt""
	""regexp""
)

type videoID struct {
	Value string
}

func newVideoID(uuid string) (*videoID, error) {

	if !isValidUUID(uuid) {
		return nil, fmt.Errorf(""UUID (%s)its not valid"", uuid)
	}

	vid := &videoID{
		Value: uuid,
	}

	return vid, nil
}

func isValidUUID(uuid string) bool {
	r := regexp.MustCompile(""^[a-fA-F0-9]{8}-[a-fA-F0-9]{4}-4[a-fA-F0-9]{3}-[8|9|aA|bB][a-fA-F0-9]{3}-[a-fA-F0-9]{12}$"")
	return r.MatchString(uuid)
}
\end{lstlisting}

\paragraph{AI-Generated}
\begin{lstlisting}[style=duolens,language=Go]
package ifname

import (
	""sync""
	""time""
)

// TTLValType represents a value with a timestamp
type TTLValType struct {
	time time.Time // when entry was added
	val  valType
}

// timeFunc is a function that returns the current time
type timeFunc func() time.Time

// TTLCache is a cache with a time to live
type TTLCache struct {
	mu            sync.RWMutex
	validDuration time.Duration
	lru           LRUCache
	now           timeFunc
}

// NewTTLCache returns a new TTL cache
func NewTTLCache(valid time.Duration, capacity uint) *TTLCache {
	return &TTLCache{
		lru:           NewLRUCache(capacity),
		validDuration: valid,
		now:           time.Now,
	}
}

// Get returns the value associated with the given key and its age
func (c *TTLCache) Get(key keyType) (valType, bool, time.Duration) {
	c.mu.RLock()
	v, ok := c.lru.Get(key)
	c.mu.RUnlock()
	if!ok {
		return valType{}, false, 0
	}
	age := c.now().Sub(v.time)
	if age < c.validDuration {
		return v.val, ok, age
	} else {
		c.mu.Lock()
		c.lru.Delete(key)
		c.mu.Unlock()
		return valType{}, false, 0
	}
}

// Put inserts or updates the value associated with the given key
func (c *TTLCache) Put(key keyType, value valType) {
	c.mu.Lock()
	v := TTLValType{
		val:  value,
		time: c.now(),
	}
	c.lru.Put(key, v)
	c.mu.Unlock()
}

// Delete removes the value associated with the given key
func (c *TTLCache) Delete(key keyType) {
	c.mu.Lock()
	c.lru.Delete(key)
	c.mu.Unlock()
}
\end{lstlisting}

\subsection{C++ Samples}
\paragraph{Human-Written}
\begin{lstlisting}[style=duolens,language=C++]
/**
 * Definition for a binary tree node.
 * struct TreeNode {
 *     int val;
 *     TreeNode *left;
 *     TreeNode *right;
 *     TreeNode(int x) : val(x), left(NULL), right(NULL) {}
 * };
 */
class Solution {
 public:
  vector<vector<string>> printTree(TreeNode* root) {
    int d = depth(root);
    int n = 0;
    for (int i = 1, cnt = 0; cnt < d; i *= 2, ++cnt) n += i;
    vector<vector<string>> res(d, vector<string>(n));
    build(root, res, 0, 0, n);
    return res;
  }

 private:
  int depth(TreeNode *root) {
    if (!root) return 0;
    return max(depth(root->left), depth(root->right)) + 1;
  }

  void build(TreeNode *root, vector<vector<string>> &res, int d, int begin,
             int end) {
    if (!root) return;
    int mid = (begin + end) / 2;
    res[d][mid] = to_string(root->val);
    build(root->left, res, d + 1, begin, mid);
    build(root->right, res, d + 1, mid + 1, end);
  }
};
\end{lstlisting}

\paragraph{AI-Generated}
\begin{lstlisting}[style=duolens,language=C++]
#include <iostream>
#include <string>
#include <cstring>
#include <sys/socket.h>
#include <netinet/in.h>
#include <arpa/inet.h>

// Define the packet struct
struct packet {
    int id;
    int sequenceNumber;
    int priority;
    char payload[1024];
};

// Function to send a packet
void sendPacket(int id, int sequenceNumber, int priority, int port, int index, int sock) {
    // Initialize the packet struct
    struct packet packet;
    packet.id = id;
    packet.sequenceNumber = sequenceNumber;
    packet.priority = priority;

    // Populate the packet with a character payload based on the index
    if (index >= 0 && index < 1024) {
        packet.payload[index] = 'A';
    }

    // Convert the packet to a string
    char packetStr[1024];
    sprintf(packetStr, ""%d %d %d %s"", packet.id, packet.sequenceNumber, packet.priority, packet.payload);

    // Send the packet using the socket-based send function
    try {
        send(sock, packetStr, strlen(packetStr), 0);
    } catch (const std::exception& e) {
        std::cerr << ""Error sending packet: "" << e.what() << std::endl;
    }
}
\end{lstlisting}

\subsection{JavaScript Samples}
\paragraph{Human-Written}
\begin{lstlisting}[style=duolens,language=JavaScript]
function inspect(value, opts) {
  const ctx = {
    budget: {},
    indentationLvl: 0,
    seen: [],
    currentDepth: 0,
    stylize: stylizeNoColor,
    showHidden: inspectDefaultOptions.showHidden,
    depth: inspectDefaultOptions.depth,
    colors: inspectDefaultOptions.colors,
    customInspect: inspectDefaultOptions.customInspect,
    showProxy: inspectDefaultOptions.showProxy,
    maxArrayLength: inspectDefaultOptions.maxArrayLength,
    breakLength: inspectDefaultOptions.breakLength,
    compact: inspectDefaultOptions.compact,
    sorted: inspectDefaultOptions.sorted,
    getters: inspectDefaultOptions.getters };
  if (arguments.length > 1) {
    if (arguments.length > 2) {
      if (arguments[2] !== undefined) {
        ctx.depth = arguments[2];
      }
      if (arguments.length > 3 && arguments[3] !== undefined) {
        ctx.colors = arguments[3];
      }
    } 
    if (typeof opts === 'boolean') {
      ctx.showHidden = opts;
    } else if (opts) {
      const optKeys = Object.keys(opts);
      for (const key of optKeys) {
        if (hasOwnProperty(inspectDefaultOptions, key) || key === 'stylize') {
          ctx[key] = opts[key];
        } else if (ctx.userOptions === undefined) {
          ctx.userOptions = opts;
        }
      }
    }
  }
  if (ctx.colors) {
    ctx.stylize = stylizeWithColor;
  }
  if (ctx.maxArrayLength === null) {
    ctx.maxArrayLength = Infinity;
  }
  return formatValue(ctx, value, 0);
}
\end{lstlisting}

\paragraph{AI-Generated}
\begin{lstlisting}[style=duolens,language=JavaScript]
module.exports = function(config) {
  config.set({
    // Base path used to resolve all patterns (e.g., files, exclude)
    basePath: '',

    // Frameworks to use (Jasmine is the testing framework)
    frameworks: ['jasmine'],

    // List of files/patterns to load in the browser
    files: [
      'dist/gainda.js', // Main project file
      'test/spec/**/*.spec.js' // Test files
    ],

    // List of files/patterns to exclude
    exclude: [],

    // Preprocessors to transform files before serving them to the browser
    preprocessors: {
      'dist/gainda.js': ['coverage'] // Optional: Add coverage preprocessing
    },

    // Test results reporter to use
    reporters: ['progress'],

    // Web server port
    port: 9876,

    // Enable/disable colors in the output (reporters and logs)
    colors: true,

    // Level of logging
    // Possible values: config.LOG_DISABLE || config.LOG_ERROR || config.LOG_WARN || config.LOG_INFO || config.LOG_DEBUG
    logLevel: config.LOG_INFO,

    // Enable/disable watching file and executing tests whenever any file changes
    autoWatch: true,

    // Start these browsers
    browsers: ['Chrome'],

    // Continuous Integration mode
    // If true, Karma captures browsers, runs the tests, and exits
    singleRun: false,

    // Concurrency level
    // How many browser instances should be started simultaneously
    concurrency: Infinity
  });
};
\end{lstlisting}

\section{Dataset Visualisations}
\subsection{Source Code Dataset} \label{subsec:sourcecode}
\begin{figure}[t]
    \centering
    \begin{subfigure}[b]{0.48\textwidth}
        \centering
        \includegraphics[width=\textwidth]{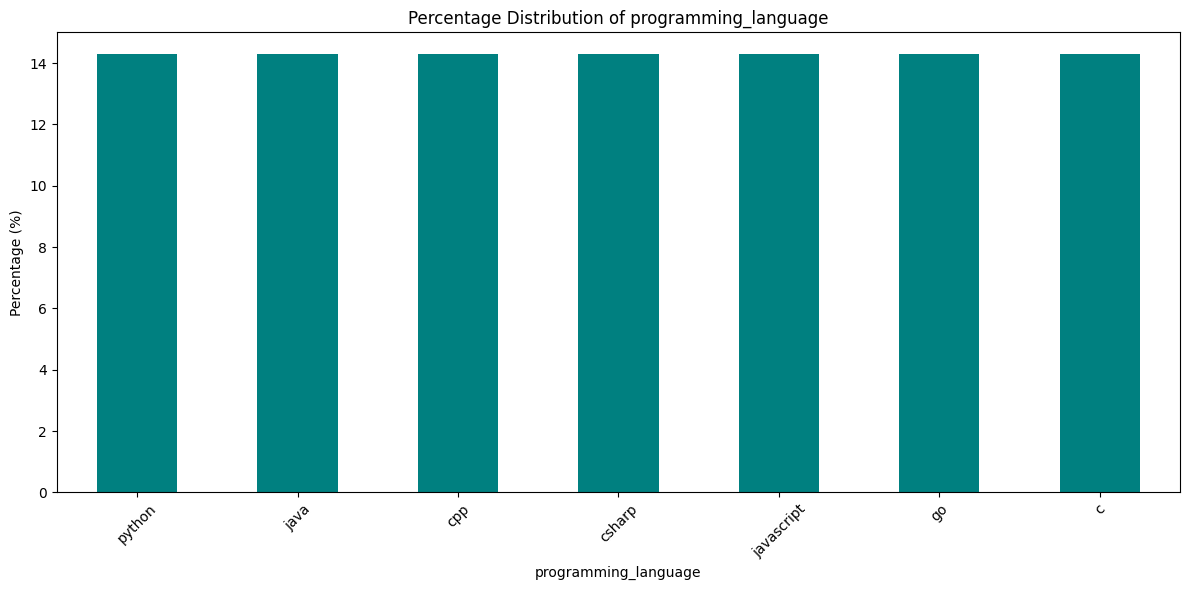}
        \caption{The number of samples for each programming language}
        \label{fig:langdistro}
    \end{subfigure}
    \hfill
    \begin{subfigure}[b]{0.48\textwidth}
        \centering
        \includegraphics[width=\textwidth]{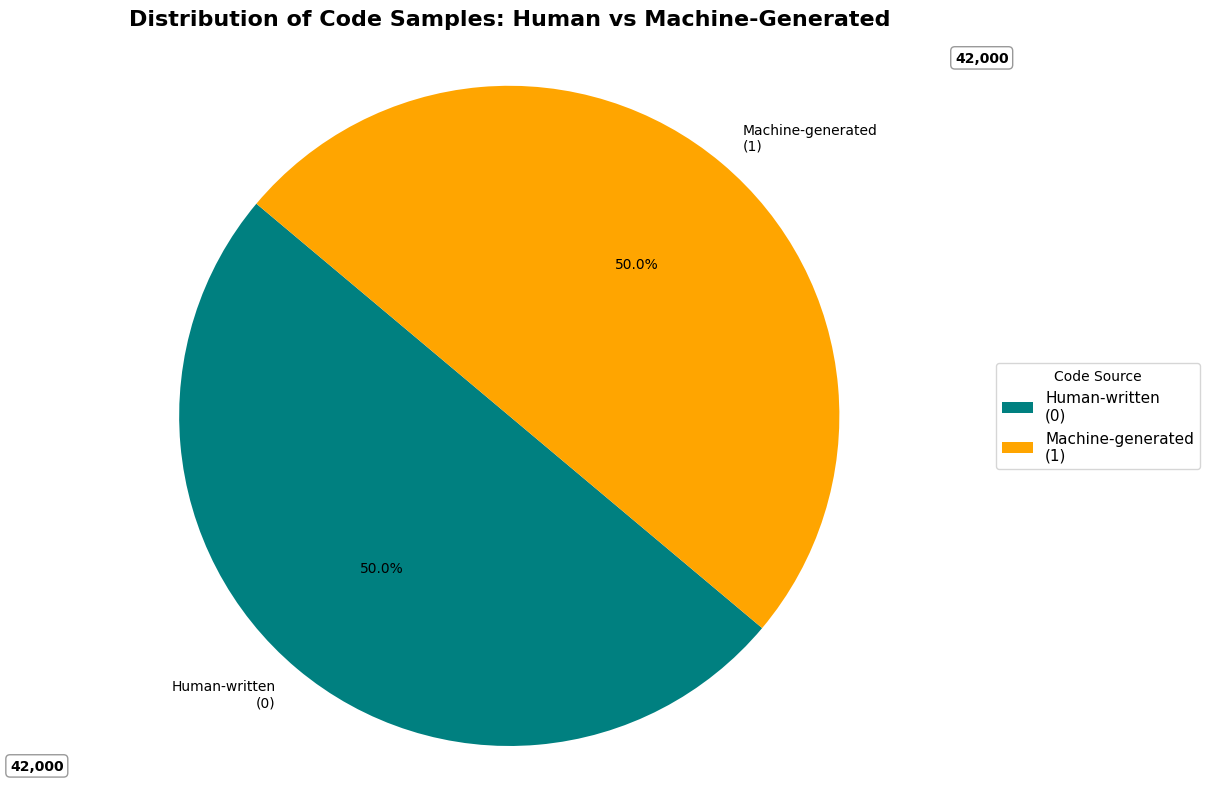}
        \caption{The percentage distribution of the samples in the dataset between human-written and machine-generated}
        \label{fig:labeldistro}
    \end{subfigure}
    
    \vspace{0.5cm}
    
    \begin{subfigure}[b]{0.48\textwidth}
        \centering
        \includegraphics[width=\textwidth]{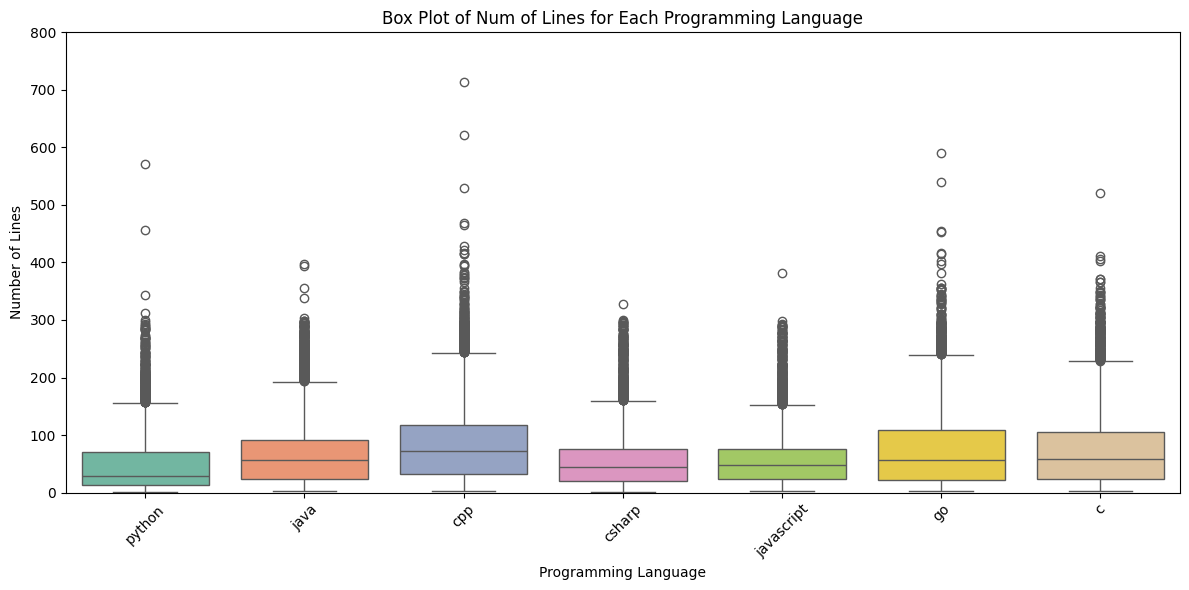}
        \caption{Box plots of the number of lines the code samples have for each language}
        \label{fig:boxplot}
    \end{subfigure}
    \caption{Source code dataset visualizations}
    \label{fig:sourcecode_viz}
\end{figure}

These visualizations illustrate how the proposed source code dataset addresses the two primary limitations observed in prior resources. As shown in Figure \ref{fig:langdistro}, the dataset maintains an equal distribution of samples across all languages, and as depicted in Figure \ref{fig:labeldistro}, it achieves a balanced representation of both human-written and machine-generated classes. This design mitigates potential biases during the fine-tuning process. Furthermore, Figure \ref{fig:boxplot} demonstrates that the dataset exhibits a substantial number of outliers with respect to sample length, which is advantageous as it reflects a broader diversity and range of examples rather than a collection of homogenous samples.

\subsection{Multilingual Text Dataset} \label{subsec:multilingual}
\begin{figure}[t]
    \centering
    \begin{subfigure}[b]{0.48\textwidth}
        \centering
        \includegraphics[width=\textwidth]{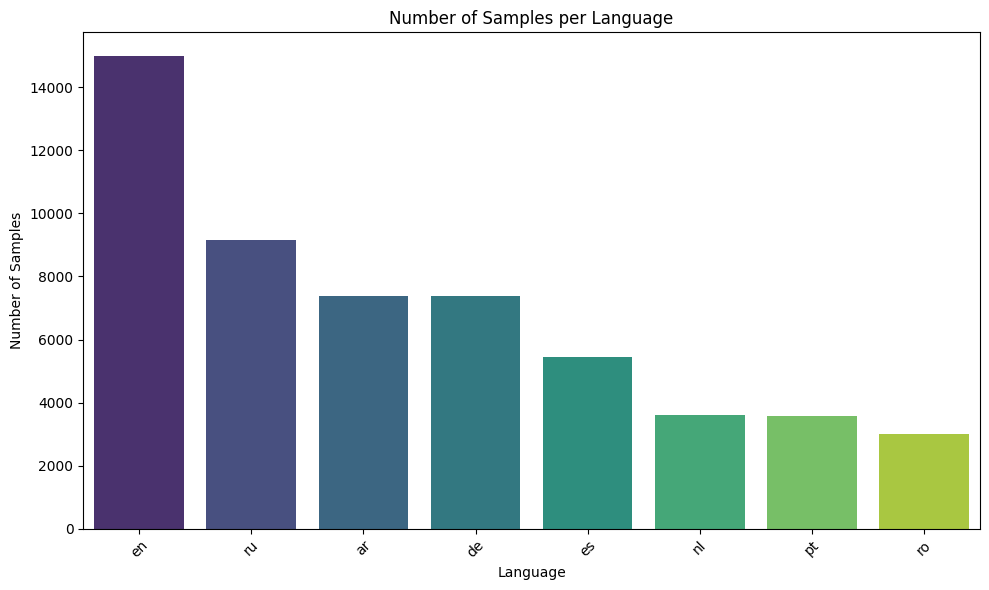}
        \caption{Number of samples per language in our dataset}
        \label{fig:multiLangDistro}
    \end{subfigure}
    \hfill
    \begin{subfigure}[b]{0.48\textwidth}
        \centering
        \includegraphics[width=\textwidth]{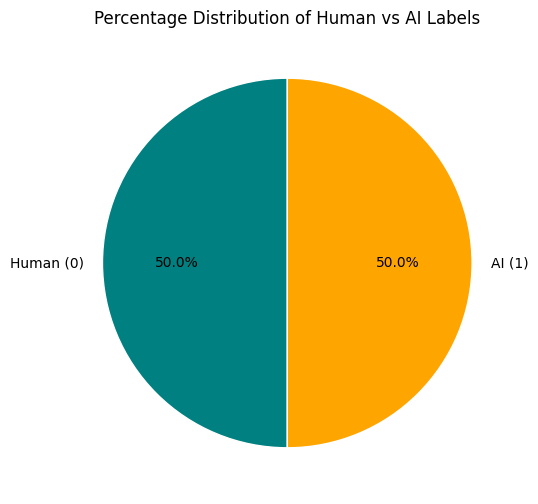}
        \caption{Percentage distribution of machine-generated and human-written samples in our dataset}
        \label{fig:multiLabelDistro}
    \end{subfigure}
    
    \vspace{0.5cm}
    
    \begin{subfigure}[b]{0.48\textwidth}
        \centering
        \includegraphics[width=\textwidth]{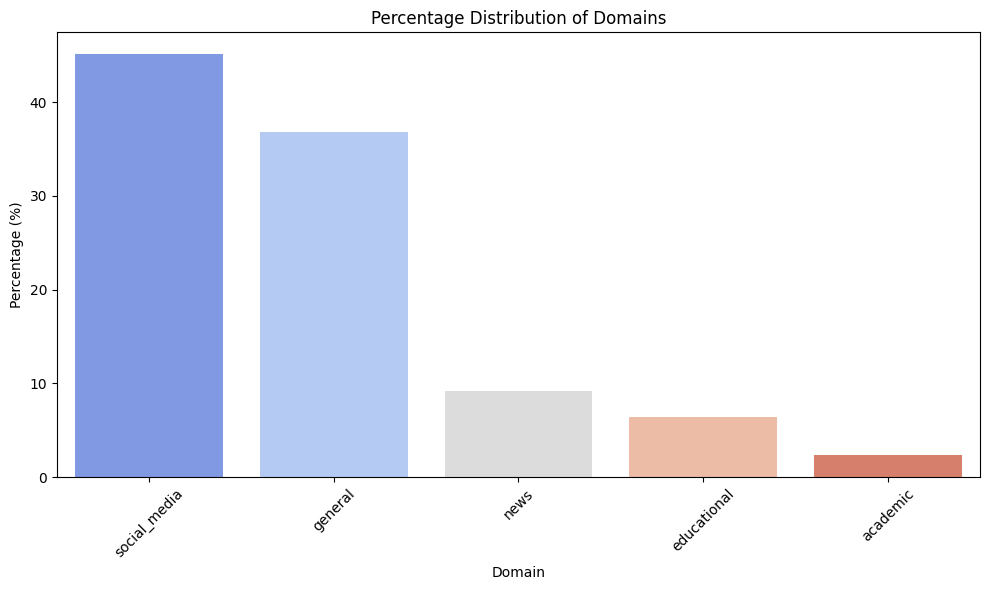}
        \caption{Percentage distribution of the domain of the samples in our dataset}
        \label{fig:multiDomainDistro}
    \end{subfigure}
    \caption{Multilingual text dataset visualizations}
    \label{fig:multilingual_viz}
\end{figure}

These visualizations demonstrate how the proposed multilingual text dataset addresses key limitations identified in prior resources. As shown in Figure \ref{fig:multiLabelDistro}, the dataset maintains a balanced distribution between human-written and machine-generated samples, thereby reducing the risk of bias during fine-tuning. In addition, Figure \ref{fig:multiDomainDistro} illustrates the balanced representation of linguistic styles: while social media samples predominantly contain casual and informal language, often including slang, the remaining domains exhibit more formal registers with comparatively refined diction. This diversity is advantageous, as it ensures broader coverage of linguistic variation while preventing overfitting to a single style. Nevertheless, the effort to simultaneously balance both domain and style necessitated a compromise in language-level balance, as reflected in Figure \ref{fig:multiLangDistro}.
\section{Language Specific Results}\label{subsec:langSpecResults}
\begin{table}[t]
    \centering
    \scriptsize
    \begin{tabular}{lcccccccl}
        \toprule
        \textbf{Model Name} & \textbf{English} & \textbf{Spanish} & \textbf{German} & \textbf{Dutch} & \textbf{Portuguese} & \textbf{Russian} & \textbf{Arabic} & \textbf{Romanian}\\
        \midrule
        DuoLens (English) & -& 0.6716& 0.7953& 0.7581& 0.7758& 0.5714 & 0.7511& 0.8606\\
        DuoLens (Spanish) & 0.6406 & -& \textbf{0.8695}& 0.8176& \textbf{0.8659}& \textbf{0.7461}& \textbf{0.8383}& 0.8726\\
        DuoLens (German)  & \textbf{0.6441}& 0.7075& -& \textbf{0.8191}& 0.8114& 0.6493& 0.8191& 0.8933\\
        DuoLens (Dutch)   & 0.6021 & 0.6911& 0.6667& -& 0.8108& 0.6134& 0.7671& 0.8836\\
        DuoLens (Portuguese)& 0.6075 & 0.7352& 0.6766& 0.7762& -& 0.6252& 0.7286& 0.88\\
        DuoLens (Russian) & 0.5917& \textbf{0.7642}& 0.8334& 0.8143& 0.8256& -& 0.7822& \textbf{0.896}\\
        DuoLens (Arabic)  & 0.6109 & 0.6512& 0.6656& 0.7501& 0.6919& 0.6496& -& 0.784\\
        DuoLens (Romanian)& 0.6053 & 0.6761& 0.6439& 0.7609& 0.8016& 0.5850& 0.7036& -\\
        \bottomrule
    \end{tabular}
    \caption{Accuracy in detection for multilingual text samples from each language}
    \label{tab:multiCrossLang}
\end{table}
\begin{table}[t]
    \centering
    \begin{tabular}{lccccccc}\toprule
         \textbf{Model Name}&\textbf{Python} &  \textbf{Java} &  \textbf{JavaScript} &  \textbf{C}&  \textbf{C\#}&  \textbf{C++}& \textbf{Go}\\\midrule
         DuoLens (Python)&-&  0.791&  0.746&  0.846&  0.775&  0.791& 0.7\\
         DuoLens (Java)&\textbf{0.649}&  -&  0.778&  \textbf{0.899}&  0.833&  \textbf{0.895}& \textbf{0.864}\\
         DuoLens (JavaScript)&0.504&  0.849&  -&  0.8&  0.865&  0.856& 0.764\\
         DuoLens (C)&0.517&  0.755&  \textbf{0.787}&  -&  0.745&  0.851& 0.792\\
         DuoLens (C\#)&0.64&  0.774&  0.711&  0.828&  -&  0.762& 0.77\\
         DuoLens (C++)&0.585&  \textbf{0.867}&  0.782&  0.884&  \textbf{0.878}&  -& 0.807\\
         DuoLens (Go)&0.467&  0.677&  0.697&  0.735&  0.69&  0.731& -\\ \bottomrule
    \end{tabular}
    \caption{Accuracy in detection for code samples from each language}
    \label{tab:crosslang}
\end{table}
\subsection{Natural Language Specific Results} 
 DuoLens was evaluated on specific languages and its accuracy per language was measured using our dataset (besides from the language it was finetuned upon). The results are included in Table \ref{tab:multiCrossLang} where the Spanish-specific DuoLens performed the best among all languages and achieved the highest accuracy out of all. 

\subsection{Programming Language Specific Results} 
 DuoLens was evaluated on specific languages and its accuracy per language was measured using our dataset (besides from the language it was finetuned upon). The results are included in Table \ref{tab:crosslang} where the Java specific DuoLens performed the best among all the languages and achieved the highest accuracy out of them. 

\section{Future Work} \label{subsec:futurework}
We identify several avenues for future work, including, but not limited to: extending coverage to additional natural and programming languages, incorporating sentence-level detection, developing agent-based or hybrid systems with LLMs to provide explanatory insights for classification outcomes, and extensions to the datasets introduced.

\end{document}